\documentclass{ws-jaic}
\usepackage{ws-jaic-thm}        
\usepackage[square]{natbib}
\usepackage{amsmath}
\usepackage{booktabs}
\usepackage{multirow}
\begin{document}

\markboth{Latapie et al., 2020}{A Metamodel and Framework for Artificial General Intelligence}

\catchline{0}{0}{0000}{}{}

\title{A Metamodel and Framework for Artificial General Intelligence \\
From Theory to Practice}

\author{Hugo Latapie, Ozkan Kilic, Gaowen Liu, Yan Yan, Ramana Kompella, Pei Wang, Kristinn R. Thórisson, Adam Lawrence, Yuhong Sun, Jayanth Srinivasa}

\address{hlatapie@cisco.com, okilic@cisco.com, gaoliu@cisco.com, yyan34@iit.edu, rkompell@cisco.com, pei.wang@temple.edu, thorisson@ru.is, adamlawr@cisco.com, yuhosun@cisco.com, jasriniv@cisco.com}

\maketitle

\pub{Received 15th December 2020}{Revised ...  December 2020}

\begin{abstract}
This paper introduces a new metamodel-based knowledge representation that significantly improves autonomous learning and adaptation. While interest in hybrid machine learning / symbolic AI systems leveraging, for example, reasoning and knowledge graphs, is gaining popularity, we find there remains a need for both a clear definition of knowledge and a metamodel to guide the creation and manipulation of knowledge. Some of the benefits of the metamodel we introduce in this paper include a solution to the symbol grounding problem, cumulative learning, and federated learning. We have applied the metamodel to problems ranging from time series analysis, computer vision, and natural language understanding and have found that the metamodel enables a wide variety of learning mechanisms ranging from machine learning, to graph network analysis and learning by reasoning engines to interoperate in a highly synergistic way. Our metamodel-based projects have consistently exhibited unprecedented accuracy, performance, and ability to generalize. This paper is inspired by the state-of-the-art approaches to AGI, recent AGI-aspiring work, the granular computing community, as well as Alfred Korzybski's general semantics. One surprising consequence of the metamodel is that it not only enables a new level of autonomous learning and optimal functioning for machine intelligences, but may also shed light on a path to better understanding how to improve human cognition.

\keywords{Artificial Intelligence \and AI \and AGI  \and General Semantics \and Levels of Abstraction \and Neurosymbolic \and Cognitive Architecture}
\end{abstract}

\section{Introduction}	
The \footnote{Preprint of an article submitted for consideration in [Journal of Artificial Intelligence and Consciousness] © [2021] [copyright World Scienti!c Publishing Company] [https://www.worldscientific.com/worldscinet/jaic]} field of artificial intelligence has advanced considerably since its inception in 1956 at the Dartmouth Conference organized by Marvin Minsky, John McCarthy, Claude Shannon, and Nathan Rochester. The exponential growth in compute power and data, along with advances in machine learning and in particular, deep learning, have resulted in remarkable pattern recognition capabilities. For example, real-time detection of pedestrians has recently achieved an average precision of over 55\% \citep{tan2019}. Natural language understanding systems are achieving superhuman performance on some tasks such as yes/no question answering \citep{wang2019a}. However, as Yan \citep{lecun2020} wrote,``trying to build intelligent machines by scaling up language models is like building high-altitude planes to go to the moon. You might beat altitude records, but going to the moon will require a completely different approach." Beyond the need to improve the accuracy of pattern recognition beyond current levels, current deep learning approaches suffer from susceptibility to adversarial attacks, a need for copious amounts of labeled training data, and an inability to meaningfully generalize. 

After over 30 years of intense effort, the AGI community has developed the theoretical underpinnings for AGI and affiliated working software systems \citep{wang2005,wang2006}. While achieving  human-level AGI is  arguably years to decades away, some of the currently available AGI subsystems are ready to be incorporated into non-profit and for-profit products and services. Some of the most promising AGI systems we have encountered are OpenNARS \citep{wang2006,wang2010}, OpenCog \citep{goertzel2009,goertzel2013}, and AERA \citep{thorisson2012}. We are collaborating with all three teams and have developed video analytics and Smart City applications that leverage both OpenCog  and OpenNARS \citep{hammer2019}.

After several years of applying these AGI technologies to complex, real-world problems in IoT, networking, and security at scale, we have encountered a few stumbling blocks largely related to real-time performance on large datasets and cumulative learning \citep{thorisson2019}. In order to progress from successful proofs of concept and demos to scalable products, we have developed the Deep Fusion Reasoning Engine (DFRE) metamodel and associated DFRE framework, which is the focus of this paper. We have used this metamodel and framework to bring together a wide array of technologies ranging from machine learning, deep learning, and probabilistic programming to the reasoning engines operating under the assumption of insufficient knowledge and resources (AIKR) \citep{wang2005}. As discussed below, we believe the initial results are promising: The data show  a dramatic increase in system accuracy, ability to generalize, resource utilization, and real-time performance when compared to state-of-the-art AI systems.

The following sections will cover related theories and technologies, the metamodel itself, empirical results, and discussions as well as future work. Appendix A contains some background information that may help readers gain a deeper understanding of this material. 

\section{Related Theories and Technologies}
\subsection{Korzybski}
After living through WWI, Korzybki, was concerned about the future trajectory of mankind. He focused his research on the creation of a non-metaphysical definition of man that was both descriptive and predictive from a scientific and engineering perspective. He focused on what he termed the ``time-binding property" that enabled human societies to advance exponentially from a technological perspective. As his objective was to discover the source of humanity's self-destructive tendencies, he created a model of the human nervous system he called ``the structural differential", which is the primary inspiration for our metamodel. Korzybski developed a theory that explains the power of the human nervous system, the weaknesses that cause many of humanity's major problems such as world wars, and a path to optimal/correct functioning of the human nervous system. The Institute of General Semantics, which Korzybski founded, continues to train of educators around the world.

Korzybski focused on helping people better utilize the considerable power of the human nervous system in part because the combination of exponential advancement of technology and a primitive way of using it could lead to large-scale destruction. Given that we have far more powerful compute capability and weapons, the sane operation of all autonomous learning systems, human or machine, is of even greater importance.

\subsection{OpenNARS}
OpenNARS (see \citep{hammer2019}) is a Java implementation of a Non-Axiomatic Reasoning System (NARS). NARS is a general-purpose reasoning system that works under the Assumption of Insufficient Knowledge and Resources (AIKR). As described in \cite{wang2009}, this means the system works in Real-Time, is always open to new input, and operates with a constant information processing ability and storage space. An important part is the Non-Axiomatic Logic (see \cite{wang2010} and \cite{wang2006}) which allows the system to deal with uncertainty.
To our knowledge, our solution is the first to apply NARS to a real-time visual reasoning task.

\subsection{Embeddings}
Graph embedding \citep{cui2018,hamilton2017} is a technique used to represent graph nodes, edges, and sub-graphs in vector space that other machine learning algorithms can use. Graph neural networks use graph embeddings to aggregate information from graph structures in non-Euclidean
ways. This allows the DFRE Framework to use the embeddings to learn from different data sources that are in the form of graphs, such as Concept Net \citep{speer2017}. Despite its performance across different domains, the graph neural networks suffer from scalability issues \citep{ying2018,zhou2018} because calculating the Laplacian matrix for all nodes in a large network may not be feasible. The levels of abstraction and the focus of attention mechanisms used by an Agent resolve these scalability issues in a systematic way.  

\section{The DFRE Metamodel}
\begin{figure}[ht]
\includegraphics[scale=0.5,width=\linewidth]{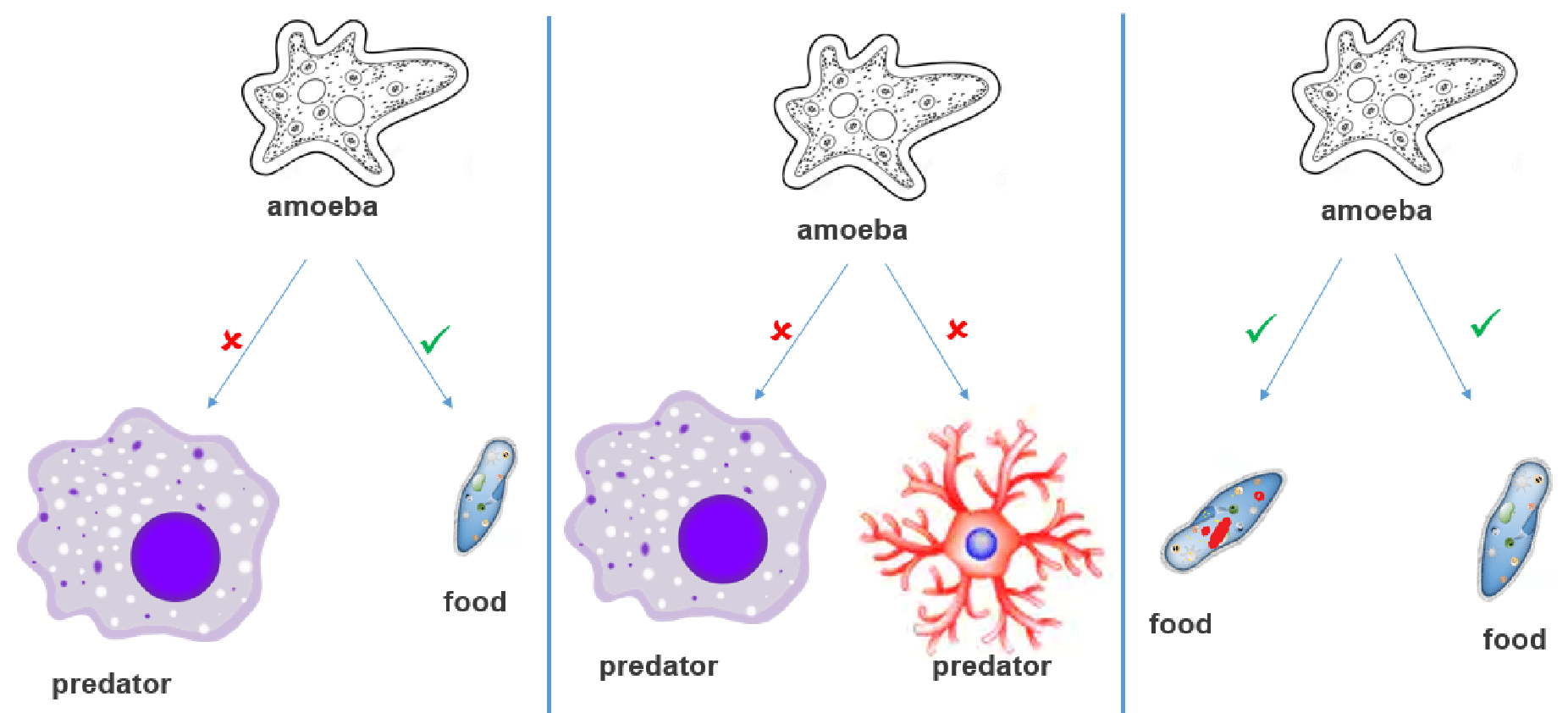}
\caption{\label{fig:amoeba}Amoeba distinguishing between distinctions and similarities.}
\end{figure}

The DFRE metamodel and framework are based on the idea that knowledge is a hierarchical structure, where the levels in the hierarchy correspond to levels of abstraction. The \textit{DFRE metamodel} refers to the way that knowledge is hierarchically structured while a \textit{model} refers to knowledge stored in a manner that complies to the DFRE metamodel. It is based on non-Aristotelian, non-elementalistic systems of thinking. The backbone of its hierarchical structure is based on \textit{difference}, a.k.a. antisymmetric relations, while the offshoots of such relations are based on symmetric relations. As in figure \ref{fig:amoeba}, even a simple amoeba has to differentiate distinctions and similarities because preserving symmetric and antisymmetric relations is fatally important.

Korzybski \citep{korzybski1994}  dedicated the majority of his professional life to analyzing and studying the nature of this hierarchical structure. While it is well beyond the scope of this paper to discuss the details of his analysis, our initial focus was to incorporate these fundamental principles.
\begin{itemize}
  \item K1 – the core framework of knowledge is based on anti-symmetric relations
        \begin{itemize}
            \item Spatial understanding: right/left/top/bottom
            \item Temporal understanding: before/after
            \item Corporal understanding: pain/satiation
            \item Emotional understanding: happy/sad
            \item Social understanding: friend/foe
            \item Causal understanding: X causes Y

        \end{itemize}
  \item K2 – symmetric relations add further structure
        \begin{itemize}
            \item A and B are friends
            \item A is like B
        \end{itemize}
  \item K3 – knowledge is layered
        \begin{itemize}
            \item Sensor data is on a different layer than high level symbolic information
            \item Symbolic information B, which expands or provides context to symbolic information A, is at a higher layer / level of abstraction
            \item In the symbolic space, there are theoretically an infinite number of layers, i.e., it is always possible to refer to a symbol and expand upon it, thus creating yet another level of abstraction
        \end{itemize}
  \item K4 – since knowledge is structure, any structure destroying operations such as confusing levels of abstraction, treating an anti-symmetric relation as symmetric, or vice-versa, can, if inadvertently applied, be knowledge-corrupting and/or a knowledge-destroying operation.\footnote[1]{Korzybski argues that what is currently limiting humanity’s advancement is the general lack of understanding of how our own abstracting mechanisms work \citep{korzybski1949}. He considers mankind to currently be in the childhood of humanity and the day, if it should come, that humanity becomes generally aware of the metamodel, is the day humanity enters into the “manhood of humanity” \citep{korzybski1921}}  However it should be noted that creative problem solving and other adaptive behaviors may require mixing levels of abstraction. The key is to ensure that the long-term structure of the metamodel is meticulously maintained and that these operations occur by design and not by accident.
\end{itemize}

\begin{figure}[h]
\includegraphics[width=\linewidth]{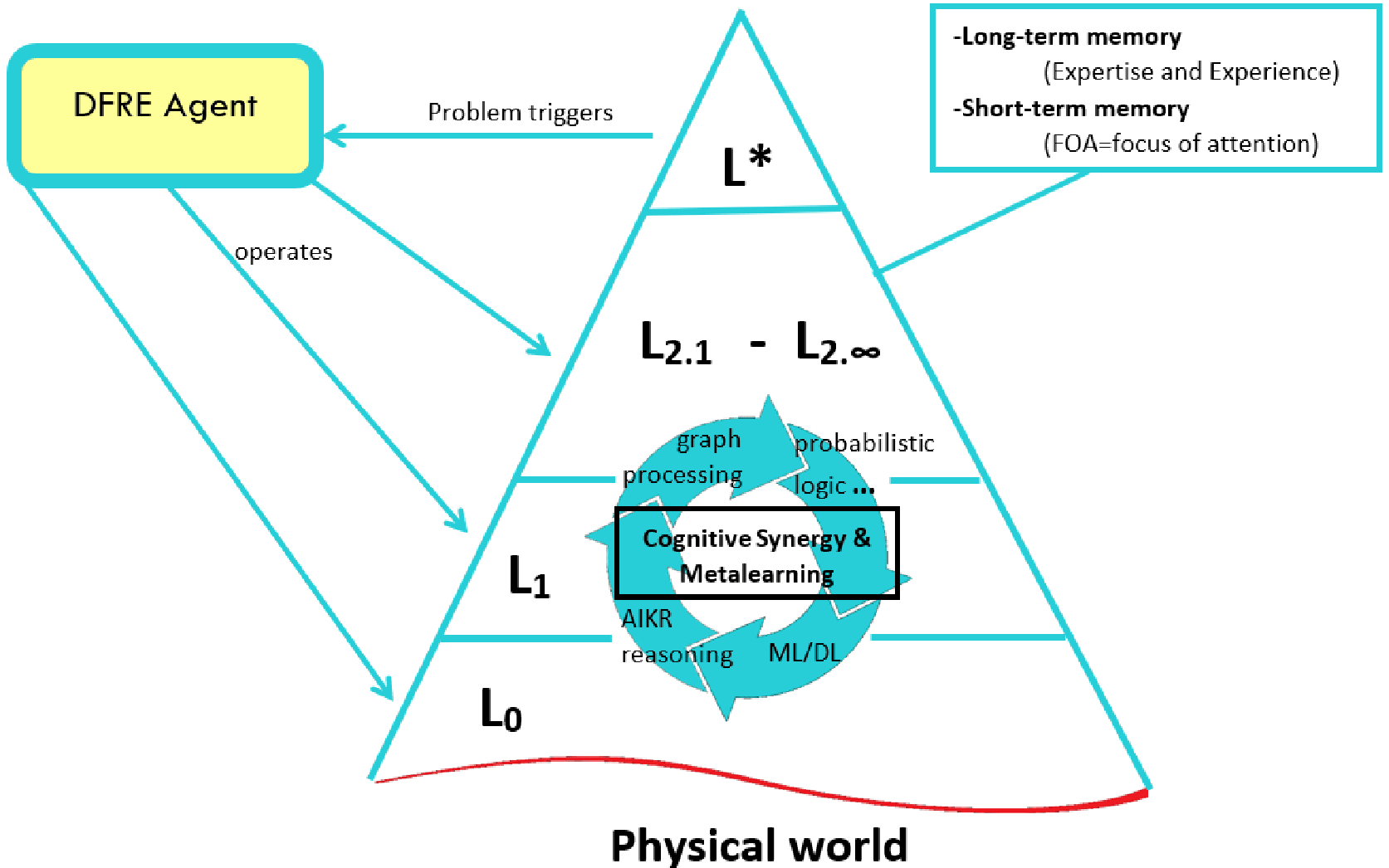}
\caption{\label{fig:loa}DFRE Framework with four levels of abstraction.}
\end{figure}

The DFRE Knowledge Graph (DFRE KG) groups information into four levels as shown in Figure \ref{fig:loa}. These are labeled L0, L1, L2, and L* and represent different levels of abstraction with L0 being closest to the raw data collected from various sensors and external systems, and L2 representing the highest levels of abstraction, typically obtained via mathematical methods, i.e. statistical learning and reasoning.  The layer L2 can theoretically have infinitely many sub-layers. L* represents the layer where the high-level goals and motivations, such as self-monitoring, self-adjusting and self-repair, are stored. There is no global absolute level for a concept and all sub-levels in L2 are relative. However, L0, L1, L2 and L* are global concepts themselves. For example, an Agent, which is basically a computer program that performs various tasks autonomously, can be instantiated to troubleshoot a problem, such as one related to object recognition or computer networking. The framework promotes cognitive synergy and metalearning, which refer to the use of different computational techniques (e.g., probabilistic programming, Machine Learning/Deep Learning, and such) to enrich its knowledge and address combinatorial explosion issues. 

\begin{figure}[h]
\includegraphics[width=\linewidth]{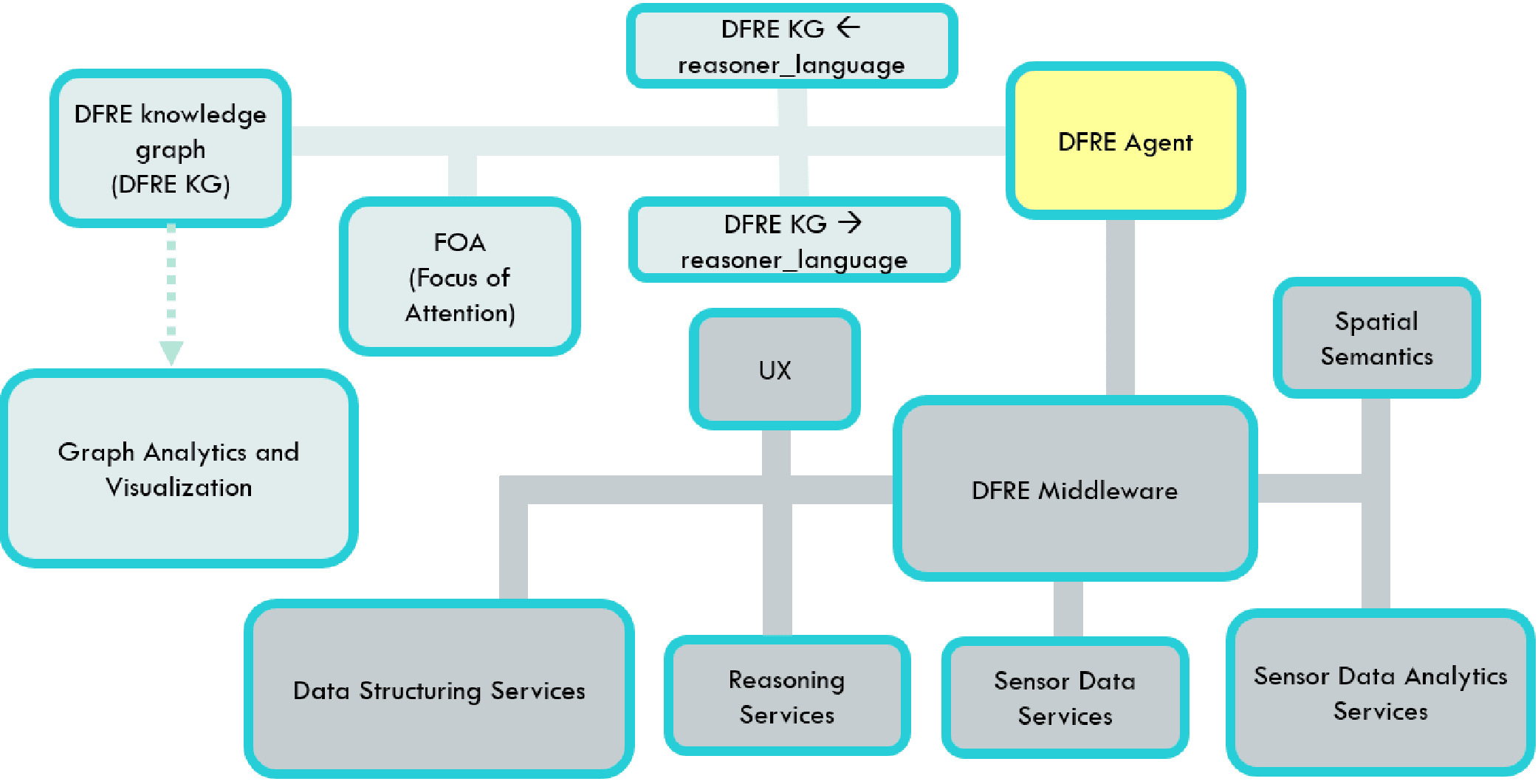}
\caption{\label{fig:architecture}DFRE Framework.}
\end{figure}

One advantage of the DFRE Framework is its integration of human domain expertise, ontologies, prior learnings by the current DFRE KG-based system and other similar systems, and additional sources of prior knowledge through the middleware services. It provides a set of services that an Agent can utilize as shown in Figure \ref{fig:architecture}. 

The Sensor Data Services are used to digitize any real world data, such as video recordings. Similarly, the Data Structuring Services restructure data ,e.g., rectifying an image, if needed. These two services are the basis for Image Processing Services which provide a set of supervised and unsupervised algorithms to detect objects, colors, lines, and other visual criteria. The Sensor Data Analytic Services analyze objects and create object boundaries enriched with local properties, such as an object's size and coordinates, which create a 2D symbolic representation of the world. Spatial Semantic Services then uses this representation to construct the initial knowledge graph that captures the spatial relations of the object as a relational graph. Any L2- or high-level reasoning is performed on this knowledge graph. 

Graph-based knowledge representation provides a system with the ability to:
\begin{itemize}
    \item Effectively capture the relations in the sub-symbolic world in a world of symbols,
    \item Keep a fluid data structure independent of programming language, in which Agents running on different platforms can share and contribute, 
    \item Use algorithms based on the graph neural networks to allow preservation of topological dependency of information \citep{scarselli2009} on nodes.
\end{itemize}

All processes are fully orchestrated by the Agent that catalogues knowledge by strictly preserving the structure while evolving new structures and levels of abstraction in its knowledge graph because, for DFRE KG, knowledge is structure. Multiple Agents can have not only individual knowledge graphs but also a single knowledge graph on which  all can cooperate and contribute. In other words, multiple Agents can work toward the same goal by sharing the same knowledge graph synchronously or asynchronously. Different Agents can have partially or fully different knowledge graphs depending on their experience, and share those entire graphs or their fragments  through the communication channel provided by the DFRE Framework. Note that although the framework can provide supervised machine learning algorithms if needed, the current IoT use case is based on a retail store which requires unsupervised methods as explained in the next section.

\section{Experimental Results}
The DFRE Framework was previously tested in the Smart City domain \citep{hammer2019}, in which the system learns directly from experience with no initial training required (one-shot), based on a fusion of sub-symbolic (object tracker) and symbolic (ontology and reasoning) information. The current use case is based on object-class recognition in a retail store. Shelf space monitoring, inventory management and alerts for potential stock shortages are crucial tasks for retailers who want to maintain an effective supply chain management. In order to expedite and automate these processes, and reduce both the requisite for human labor and the risk of human error, several machine learning and deep learning-based techniques have been utilized \citep{baz2016,franco2017,george2014,tonioni2017}. Despite the high success rates, the main problems for such systems are the requirements for a broad training set, including compiling images of the same product with different lighting and from different angles, and retraining when a new product is introduced or an existing product is visually updated.  The current use case does not demonstrate an artificial neural network-based learning. The DFRE Framework has an artificial general intelligence-based approach to these problems.

\begin{figure*}[h]
\includegraphics[width=\linewidth]{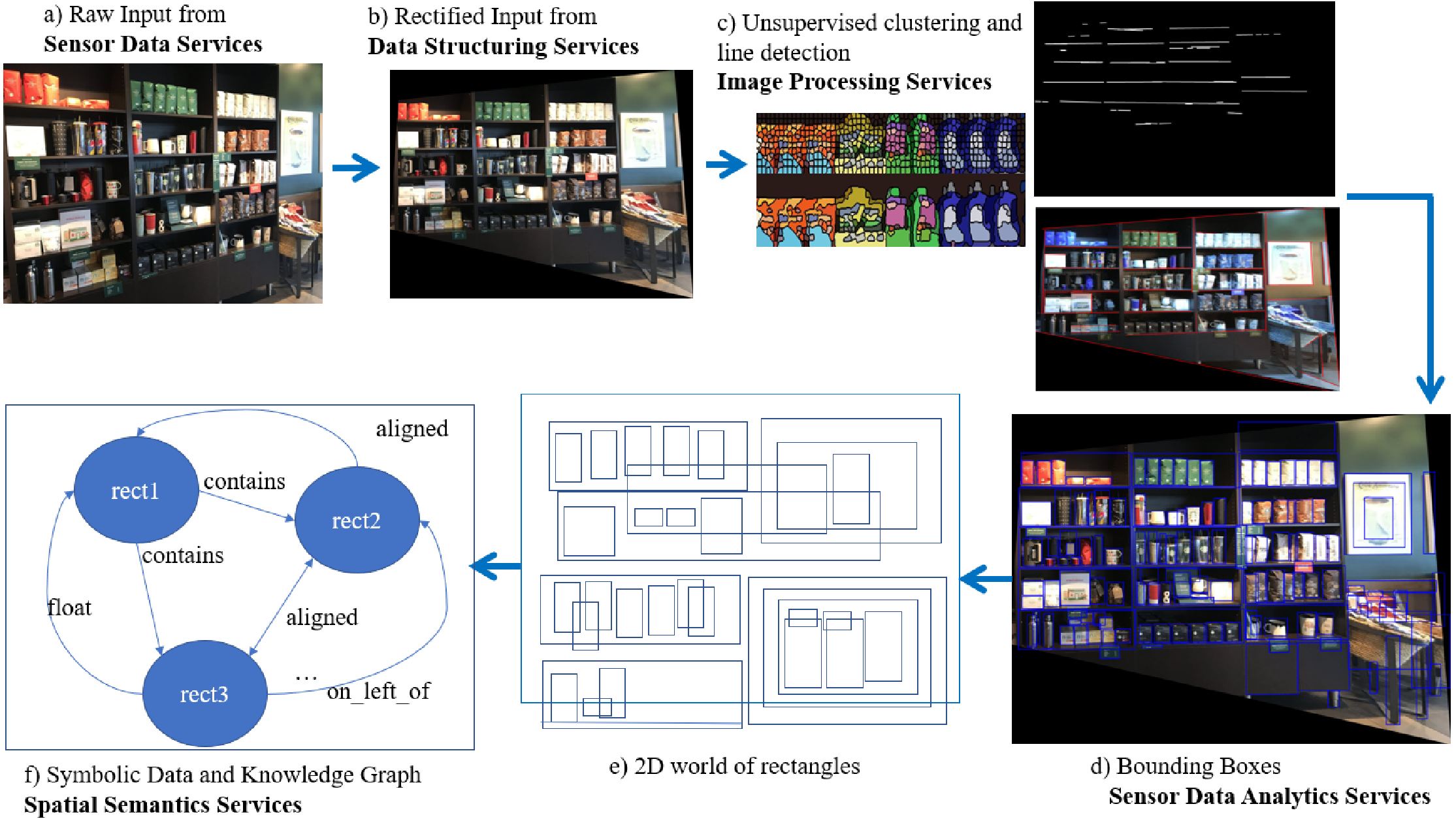}
\caption{\label{fig:retail} Retail use case for DFRE Framework.} 
\end{figure*}

Before a reasoning engine operates on symbolic data within the context of the DFRE Framework, several services must be run, as shown in Figure \ref{fig:retail}.   The flow starts with a still image captured from a video camera that constantly records the retail shelves by the Sensor Data Services as in Figure \ref{fig:retail}.a, which corresponds to L0 in Figure \ref{fig:loa}. Next, the image is rectified by the Data Structuring Services in  Figure \ref{fig:retail}.b for better line detection by the Image Processing Services, as displayed in  Figure \ref{fig:retail}.c. The Image Processing Services in the retail case are unsupervised algorithms used for color-based pixel clustering and line detection, such as probabilistic Hough transform \citep{kiryati1991}. The Sensor Data Analytics Services in Figure \ref{fig:retail}.d create the bounding boxes which represent the input in a 2D world of rectangles, as shown in Figure \ref{fig:retail}.e. The sole aim of all these services is to provide the DFRE KG with the best symbolic representation of the sub-symbolic world in rectangles. Finally, the Spatial Semantics Services operate on the rectangles to construct a knowledge graph, which preserves not only the symbolic representation of the world, but also the structures within it in terms of relations, as shown in  Figure \ref{fig:retail}.f. This constitutes the L1 level abstraction in the DFRE KG. L1 knowledge graph representation also recognizes and preserves the attributes of each bounding box, such as the top-left \textit{x} and \textit{y} coordinates, and the \textit{center}'s coordinates: \textit{height}, \textit{width}, \textit{area} and \textit{circumference}. The relations used for the current use case are \textit{inside}, \textit{aligned}, \textit{contains}, \textit{above}, \textit{below}, \textit{on left of}, \textit{on right of}, \textit{on top of}, \textit{under} and \textit{floating}. Since the relations in the DFRE KG are by default antisymmetrical, the system does not know that \textit{aligned(a,b)} means \textit{aligned(b,a)}, or \textit{on left of} and \textit{on right of} are inverse relations unless such terms are input as expert knowledge or are learned by the system through experience or simulations. The only innate relations in the DFRE metamodel are \textit{distinctions}, which are \textit{anti-symmetric} and \textit{similarity} relations; and the rest is learned by experience.

\begin{figure}[h]
    \includegraphics[width=\linewidth]{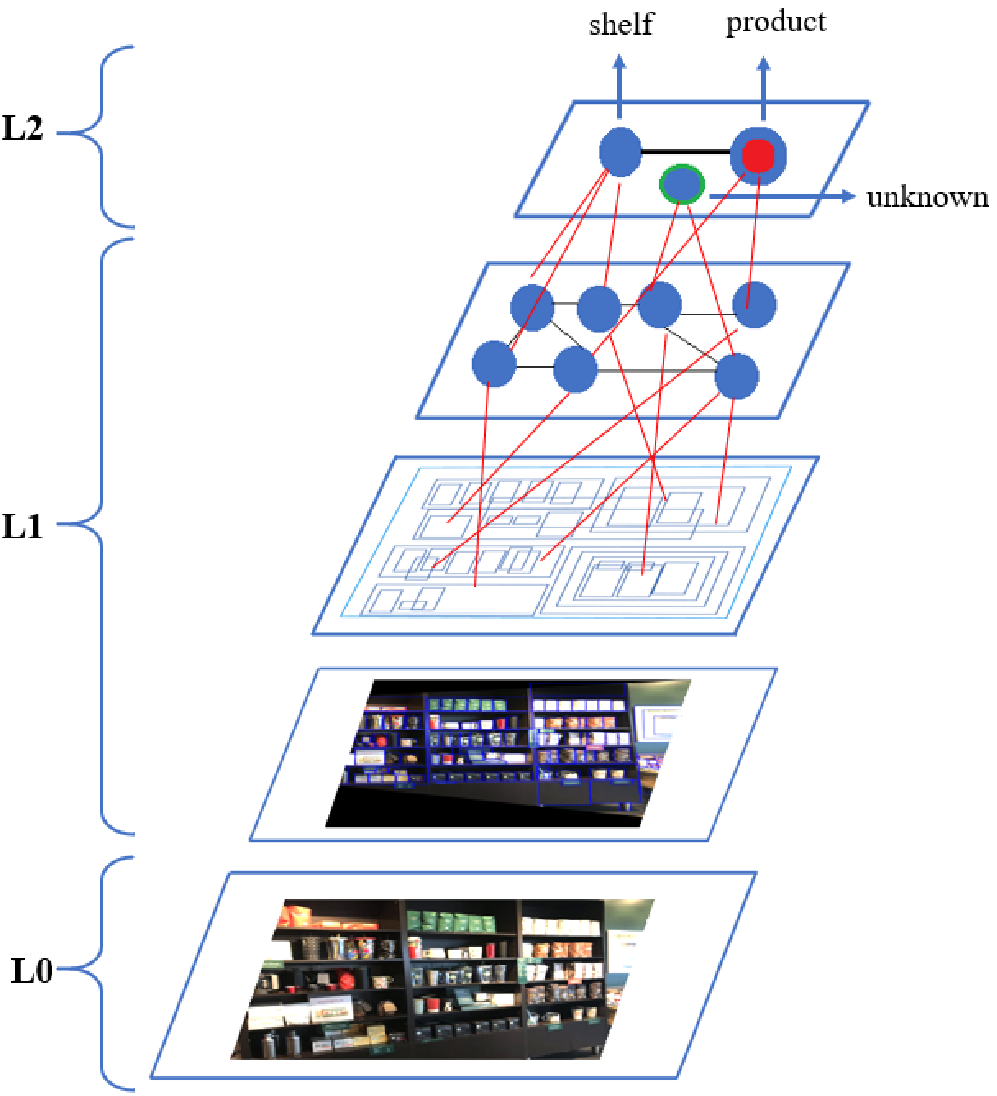}
    \caption{\label{fig:loaretail}LoA for retail use case.}
\end{figure}
The system's ultimate aim is to dynamically determine \textit{shelves}, \textit{products} and \textit{unknown/others}, as illustrated in Figure \ref{fig:loaretail}, and to monitor the results with timestamps.  

While L2 identifies only the concepts of \textit{shelf}, \textit{product} and \textit{unknown}, and the possible relations among them, the reasoning engine, NARS \citep{wang2006,wang2010,wang2018}, creates their L1 intensions as an evidence-based truth system in which there is no absolute knowledge. This is useful in the retail use case scenario because the noise in L0 data causes both overlapping regions and conflicting premises at L1. This noise results from not only the projection of the 3D world input data into a 2D framework, but also the unsupervised algorithms used by L1 services. The system has only four rules for L2 level reasoning:
\begin{itemize}
            \item \textit{If a rectangle contains another rectangle that is not floating, the outer rectangle can be a shelf while the inner one can be a product.} 
            \item \textit{If a rectangle is aligned with a shelf, it can be a shelf too.}
            \item \textit{If a rectangle is aligned with a product horizontally, it can be a product too.}
            \item \textit{If a floating rectangle is stacked on a product, it can be a product too.}
\end{itemize}
Note that applying levels of abstraction gives the DFRE Framework the power to perform reasoning based on the expert knowledge in L2 level mostly independent of L1 level knowledge. In other words, the system does not need to be trained for different input; it is unsupervised in that sense. The system has a metalearning objective which continuously attempts to improve its knowledge representation. The current use case had 152 rectangles of various shapes and locations, of which 107 were products, 16 were shelves, and the remaining 29 were other objects. When the knowledge graph in L1 is converted into Narsese, 1,478 lines of premises that represent both the relations and attributes were obtained and sent to the reasoner. Such a large amount of input with the conflicting evidence caused the reasoning engine to perform poorly. Furthermore, the symmetry and transitivity properties associated with the reasoner resulted in the scrambling of the existing structure in the knowledge graph. Therefore, the DFRE Framework employed a Focus of Attention (FoA) mechanism. The FoA creates overlapping covers of knowledge graphs for the reasoner to work on this limited context. Later, the framework combines the results from the covers to finally determine the intensional category. For example, when the FoA utilizes the reasoner on a region, a rectangle can be recognized as a shelf. However, when the same rectangle is processed in another cover, it may be classified as a product. The result with higher frequency and confidence wins. The FoA mechanism is inspired by the human visual attention system, which manages input flow and recollects evidence as needed in case of a conflicting reasoning result. A FoA mechanism can be based on objects' attributes, such as color or size, with awareness of proximity. For this use case, the FoA determined the contexts by picking the largest non-empty rectangle, and traversing its neighbors based on their sizes in decreasing order. 

\begin{figure*}[h]
\includegraphics[width=\linewidth]{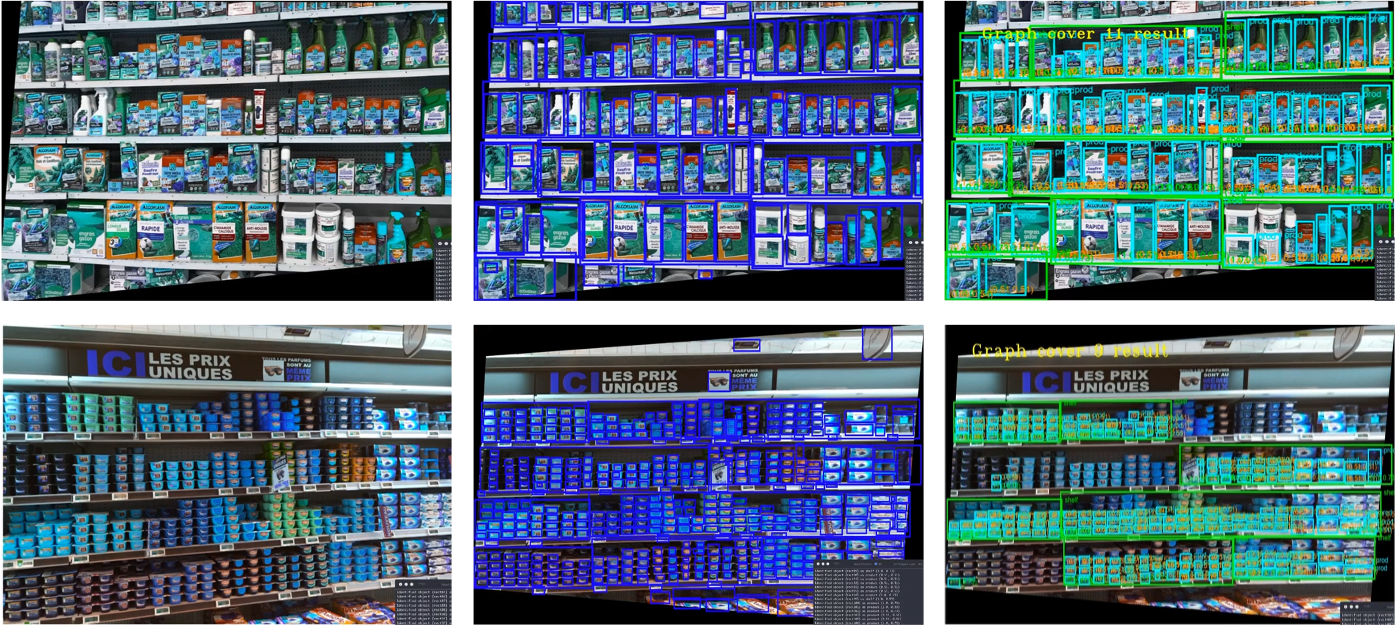}
\caption{\label{fig:visual_experiments} DFRE Framework qualitative result in retails use cases with different environments} 
\end{figure*}

The framework is tested in various settings with different camera angles and products placements as shown in Figure \ref{fig:visual_experiments}.

In Figure \ref{fig:visual_experiments}, each row represents samples from different settings: rectified frames, bounding boxes, and instantaneous output of the reasoner. We would like to emphasize that the system does not require any retraining or any change in order to adapt to the new setting. It requires only a camera to be pointed to the scene; then it automatically generalizes.

DFRE Framework was tested 10 times in 4 different settings with and without the FoA. The precision, recall and f-1 scores are exhibited in Table \ref{tab:result}.

\begin{table*}[h]
    \centering
    \caption{\label{tab:result}DFRE Framework experimental results.}
    \resizebox{0.97\linewidth}{!}{%
    \begin{tabular}{cccccll}
    \cline{1-7}
     \multirow{2}{*}{\textbf{Category}} & \multicolumn{3}{c}{\textbf{without FoA }(\%)} & \multicolumn{3}{c}{\textbf{with FoA } (\%)}\\ 
           & precision      & recall     & f1-score    & precision    & recall   & f1-score    \\ \hline
    \multicolumn{1}{c}{\textit{product}}          & 80.70   & 29.32   & 52.88   & 96.36   & 99.07   & 97.70           \\
    \multicolumn{1}{c}{\textit{shelf}}            & 8.82    & 18.75   & 12.00   & 82.35   & 87.50   & 88.85           \\
    \multicolumn{1}{c}{\textit{other}}            & 36.61   & 89.66   & 52.00   &96.00    & 82.76  & 88.89          \\ \hline
    \multicolumn{1}{c}{\textbf{overall accuracy}} & \multicolumn{3}{c}{\textbf{46.30 } \newline (min/max: 30.13/84.65)}           & \multicolumn{3}{c}{\textbf{94.73 } \newline (min/max: 88.10/100.00)}        \\ \hline
    \end{tabular}}
\end{table*}

The results indicate that the FoA mechanism improves the success of our AGI-based framework significantly by allowing the reasoner to utilize all of its computing resources in a limited but controlled context. The results are accumulated by the framework, and the reasoner makes a final decision. This approach not only allows us to perform reasoning on the intension sets of L1 knowledge, which are retrieved through unsupervised methods, but also resolves the combinatorial explosion problem whose threshold depends on the limits of available resources. In addition, one can easily extend this retail use case to include prior knowledge of product types and other visual objects, such as tables, chairs, people and shelves, as allowed by the DFRE KG.  

\subsection{Graph Embedding for Link Predictions}

Recall, a graph $G(V,E)$, where $V$ is the set of all vertices, or nodes, in $G$, and $E$, is the set of paired nodes, called edges. $|V| \in \mathbb{Z}$ is the order of the graph, or the total number of nodes. 

As mentioned before, DFRE takes advantage of graph embedding, a transformation that constructs a non-dimensional knowledge graph $G$ into a $d$-dimensional vectors space $S \in \mathbb{R}^{ |V| \times d}$. Among the many benefits, such as creating a Euclidean distance measurement for $G$, link predictions can be established between node vectors in $S$. Preliminary experimental results have given a great deal of insight into the relationships between nodes that might otherwise not be present from the graph space. 

The main algorithm used by DFRE to transform our knowledge graph $G$ to a 2-dimensional vector space is Node2Vec. \citep{grover2016}. This framework is a representation learning based approach that learns continuous feature representations for all nodes in a given knowledge graph $G$. The benefits from this algorithm, and the motivation for use in DFRE, construct the graph embedding space $S$ where link prediction, and other methods of measurement, can be used while preserving relevant network properties from the original knowledge graph. 

The functionality behind Node2Vec is similar to most other embedding processes, by use of the Skip-Gram model, and a sampling-strategy. Four arguments are input into the framework: the number of walks, the length of the walks, $p$, and $q$, where $p$ is referred to as the return hyper-parameter, and $q$ is the I/O hyper-parameter.

Once a 2-dimensional vector representation has been assigned to every node $ n \in V $ , our embedding vectors space $S \in \mathbb{R}^{ |V| \times 2}$ can provide additional metrics used for machine learning and prediction measures. One such measure is link prediction used to understand the relationship between nodes in a graph that might not be obvious from the graph space. 

Consider the nodes $n_1 , n_2 \in G(V,E)$ such that $n_1$ and $n_2$ are not similar ideas in the graph (e.g. the probability $(n_1 , n_2) \in E(G)$ is low). Once the nodes are represented in vector form $\hat{n_1} , \hat{n_2} \in \mathbb{R}^{2} \subset S$, we establish a linear relationship between the two such that a line $ y = ax + b $ is satisfied, where $a = \frac{\hat{n_{22}} - \hat{n_{21}}}{\hat{n_{12}} - \hat{n_{11}}}$ and $b = \hat{n_{21}} - a(\hat{n_{11}})$. 

Let $ \epsilon > 0 $ , then  $\forall \hat{n_k}$ that lies within the range of $ y \pm \epsilon$ , we consider these node vectors to be associated hidden links between two the two ideas $\hat{n_1}$ and $\hat{n_2}$. 

Additionally, if the line $y$ is divited into four evenly distributed quadrants ${y_1 .. y_4}$ and grown by small perturbations where $ y^{\prime} = y \pm \epsilon^{\prime} $ such that $\epsilon^{\prime} = \epsilon + \gamma$ and $\gamma \in (0,1]$ until there exists at least one $\hat{n_k}$ in every quadrant. We call this set of node vectors $S_{n}$.

This set of node vectors $S_n$ gathered within range $y^{\prime}$ provide DFRE a relationship that might not be immediately obvious from the graph space alone. The distribution of the vectors along the quadrants is revealing in away such that, for example, consider the two disjoint subsets $(\hat{n_k})_i$ and $(\hat{n_l})_j$ of $S_n$. Without loss of generality, if $(\hat{n_k})_i \in y^{\prime}_1$ and $(\hat{n_k})_j \in y^{\prime}_4$, where $ i << j $, we see that the relationship skews towards the set of node vectors that lie within the range of $y_4^{\prime}$.

Additionally, within the embedding space, consider a finite set of clusters $\{C_1 , C_2 , \dots \}$, each corresponding to its own central idea. For any arbitrary cluster $C_i$, if a new node vector ${\hat{n^{\prime}}}$ is introduced in $S$ such that ${\hat{n^{\prime}}} \in C_i$, then we can easily leverage this proximity into our sub-symbolic space to identify any additional node vectors.   

We find the main benefit to graph embedding is that we now have an unsupervised method for correlating the graph embedding space with additional embedding spaces that are generated using unsupervised machine learning techniques.

\section{Philosophical Implications}
The nativism-versus-empiricism debate, which posits that some knowledge is innate and some is learned through experience, was ascribed in the ancient world by the Greek philosophers, including Plato and Epicurus. Today, Descartes is widely accepted as a pioneering philosopher working on the mind as he furthered and reformulated the debate  in the 17\textsuperscript{th} century with new arguments. Perception, memory, and reasoning are three fundamental cognitive faculties that enhance this debate by explicating the building blocks of natural intelligence. We perceive the sub-symbolic world, and abstract it in memory, and reason on this symbolic world representation. All three place concept learning and categorization at the center of the human mind. 

The process of concept learning and categorization continues to be an active research topic related to the human mind since it is essential to natural intelligence \citep{lakoff1984} and cognitively inspired robotics research \citep{chella2006,lieto2017}. It is widely accepted that this process is based more on interactional properties and relationships among Agents, as well as between an Agent and its environment, than objective features such as color, shape and size \citep{johnson1987,lakoff1984}. This makes the distinction of anti-symmetric and symmetric relations crucial in the DFRE Framework, which assumes that the levels of abstraction are part of innate knowledge. In other words, an Agent has L0, L1 and pre-existing L2 by default. This constitutes a common a priori metamodel shared by all DFRE Agents. Each Agent instantiated from the framework has the abstraction skill based on interactional features and relationships. If the concepts in real life exist in interactional systems, natural intelligence needs to capture these systems of interactions with its own tools, such as abstraction.  These tools should also be based on interactional features by strictly preserving the distinction between symmetry and anti-symmetry. 

The mind is a system as well. Modern cognitive psychologists agree that concepts and their relations in memory function as the fundamental data structures to higher level system operations, such as problem solving, planning, reasoning and language. Concepts are abstractions that have evolved  from a conceptual primitive. An ideal candidate for a conceptual primitive would be something that is a step away from a sensorimotor experience \citep{gardenfors2000}, but is still an abstraction of experience \citep{cohen1997}. For example, a dog fails the mirror test but exhibits intelligence when olfactory skills are needed to complete a task \citep{horowitz2017}. A baby’s mouthing behavior is not only a requisite for developing oral skills but also for discovering the surrounding environment through one of its expert sensorimotor skills related to its survival. The baby is probably abstracting many objects into edible versus inedible higher categories given its insufficient knowledge and resources. What is astonishing about a natural intelligence system is that it does not need a plethora of training input and experiments to learn the abstraction. It quickly and automatically fits new information into an existing abstraction or evolves it into a new one, if needed. This is nature’s way of managing combinatorial explosion. Objects and their interconnected relationships within the world can be chaotic. Natural intelligence’s solution to this problem becomes its strength: context. The concept of ‘sand’ has different abstractions depending on whether it is on a beach, on a camera, or on leaves. An Agent in these three different contexts must abstract the sand in relation to its interaction with world in its short-term memory. This cumulative set of experiences can later become part of long-term memory, more specifically, episodic memory. The DFRE Framework uses a Focus of Attention (FoA) mechanism that provides the context while addressing the combinatorial explosion problem. The DFRE metamodel's  new way of representing practically all knowledge as temporally evolving (i.e. time series) can be viewed as the metamodel's conceptual space. For example, the retail use case given in Section 2 starts with a 3D world of pixels that is abstracted as lines and rectangles in 2D. The framework produces spatial semantics using the rectangles in the 2D world. Based on this situation, a few hundred rectangles produce thousands of semantic relations, which present a combinatorial explosion for most AGI reasoning engines. For each scene, the DFRE KG creates contexts, such candidate shelves, runs reasoners for each context, and merges knowledge in an incremental way. This not only addresses the combinatorial explosion issue, but also increases the success rate of reasoning, provided that the levels of abstraction are computed properly \citep{gorban2018}.

Abstracting concepts in relation to their contexts also allows a natural intelligence to perform mental experiments, which is a crucial part of planning and problem solving. The DFRE Framework can integrate with various simulators, re-run a previous example together with its context, and alter what is known for the purposes of  experimentation to gain new knowledge, which is relationships and interactions of concepts. 

Having granular structures provides structured thinking, structured problem solving, and structured information processing \citep{yao2012}. The DFRE Framework has granular structures but emphasizes the preservation of structures in knowledge. When a genuine problem that cannot be solved by the current knowledge arises, it requires scrambling the structures and running simulations on the new structures in order to provide an Agent with creativity. Note that this knowledge scrambling is performed in a separate sandbox. The DFRE Framework ensures that the primary DFRE KG is not corrupted by these creative, synthetic, ``knowledge-scrambling" activities.

\section{Discussion on Metamodel and Consciousness}
For the purposes of this discussion, we will define consciousness as autonomous self-aware adaptation to the environment. This means that an abstraction of the self as well as the environment of the self is learned autonomously. Human consciousness builds on the prior capabilities of chemistry-binders (plants) and space-binders (animals) with the unique ability of infinite levels of abstraction \citep{korzybski1994}. For any concept, one can envision creating a higher-level meta-concept. We are able to formulate symbolic representations that can be externalized and shared. The human model of the self evolves not only via direct interaction with the environment and cogitation, but also by watching other humans and modeling them. Human consciousness as an implementation of the metamodel appears to be dynamic in nature. The concept of self can grow to encompass family, friends, social/work groups, and beyond. Understanding our nature as time-binders that form a collective consciousness of ever-increasing power (cognitive and physical) over time, civilizations, and generations, appears to lead to higher cognitive functioning of the individual.

Human societies consisting of billions of people networked together in real-time, with petabytes of shared storage and petaflops of compute, may see the evolution of exo-cortical consciousness. In fact, many argue that this exo-cortical consciousness already exists with the growing number of autonomous self-healing systems deployed and connected throughout the world.

Since the metamodel hypothesizes the existence of an exo-cortical consciousness, it consequently yields to the possibility of implementing artificial consciousness, e.g., in robots. Artificial consciousness, which is also known as machine consciousness, is a field designed to mimic the aspects of human cognition that are related to human consciousness \citep{aleksander2008, chella2009}. In the 1950s, consciousness was seen as a vague term, and inseparable from intelligence. \citep{searle1992,chalmers1996}. Fortunately, the improvements in technology, and computational and cognitive sciences have created new interest in the field. \citep{chella2011} reviews that the most important gap between artificial and biological consciousness studies is engineering autonomy, semantic capabilities, intentionality, self-motivation, resilience and information integration. The Agents based on the metamodel have autonomy. They also have semantic capabilities and intentions to seek solutions or communicate with other Agents for knowledge sharing, which are set by self-motivation.  

\cite{chella2011} emphasize that consciousness is a real physical phenomenon, can be artificially replicated, is either a computational phenomenon or more. We bring forth the metamodel as an enabler for the achievement of artificial consciousness. The abstraction mechanism constantly and automatically creates the abstractions of the sensor data and the system's own experience. The metamodel is based on generation of new knowledge using self-perception and experience, and shares knowledge among the Agents of similar nature to support collective consciousness and resilience. The creation of self through experience gives the metamodel the ability of enhanced generalization and autonomy. Similarly, focus of the attention mechanism to segment complex problems semantically is also related to consciousness because attention and consciousness are interrelated \citep{taylor2007,taylor2009}. Implementation of attention is important because control theory is related to consciousness and plays a leading role in the intentional mechanism of an Agent.  

\section{Conclusion}
Several mathematical models and formal semantics \citep{duntsch2002,belohlavek2004,wang2008,wille1982,ma2007} are proposed to specify the meanings of real world objects as concept structures and lattices. However, they are computationally expensive \citep{jinhai2015}. One way to overcome this issue is with granular computing \citep{yao2012}. The extension of a concept can be considered a granule, and the intension of the concept is the description of the granule. Assuming that concepts share granular common parts with varying derivational and compositional stages, categorization, abstraction and approximation occur at multiple levels of granularity which plays an important role in human perception \citep{hobbs1985,yao2001,yao2009}. The DFRE Framework has granular structures but emphasizes the preservation of structures in knowledge. Being in the extension of a concept does not necessarily grant the granular concept the right to have similar relations and interactions of its intensional concept up to a certain degree or probability. Each level must preserve its inter-concept relationships and its symmetry or anti-symmetry in a hierarchically structured way.

We have outlined the fundamental principles of the DFRE Framework. DFRE takes a neurosymbolic approach leveraging state-of-the-art subsymbolic algorithms (e.g. ML/DL/Matrix Profile) and state-of-the-art symbolic processing (e.g. reasoning, probabilistic programming, and graph analysis) in a synergistic way. The DFRE metamodel can be thought of as a knowledge graph with some additional structure, which includes both a formalized means of handling anti-symmetric and symmetric relations, as well as a model of abstraction. This additional structure enables DFRE-based systems to maintain the structure of knowledge and seamlessly support cumulative and distributed learning. Although this paper provides highlights of one experiment in the visual domain employing an unsupervised approach, we have also run similar experiments on time series and natural language data with similar promising results. 

\section{Future work} 
Nowadays there is a rapid transition in the AI research field from single modality tasks, such as image classification and machine translation, to more challenging tasks that involve multiple modalities of data and subtle reasoning, such as visual question answering (VQA) \citep{agrawal2015, anderson2017,zhu2020} and visual dialog \citep{vishvak2019}. A meaningful and informative conversation, either between human-computer or computer-computer, is an appropriate task to demonstrate such a reasoning process given the complex information exchange mechanism during the dialog. 
However, most existing research focuses on the dialog itself and involves only a single Agent. We plan to design a more reliable DFRE system with implicit information sources. To this end, we propose a novel natural and challenging task with implicit information sources: describe an unseen video mainly based on the dialog between two cooperative Agents.

The entire process can be described in three phases: 
In the preparation phase, two Agents are provided with different information. Agent A1 is able to see the complete information from different modalities (i.e., video, audio, text), while Agent A2 is only given limited information. In the second phase, A2 has several opportunities to ask A1 relevant questions about the video, such as the person involved, the event happened, \textit{etc.} A2 is encouraged to ask questions that help to accomplish the ultimate video description objective, and A1 is expected to give informative and constructive answers that not only provide the needed information but also motivate A2 to ask additional useful questions in the next conversation round. After several rounds of question-answer interactions, A2 is asked to describe the unseen video based on the limited information and the dialog history with A1. In this task setup, our DFRE system  accomplishes a multi-modal task even without direct access to the original information, but learns to filter and extract useful information from a less sensitive information source, \textit{i.e.}, the dialog. It is highly difficult for AI systems to identify people based on the natural language descriptions. Therefore, such task settings and reasoning ability based on implicit information sources have great potential to be applied in a wide practical context, such as the smart hospital systems, improving current systems.

The key aspect to consider in this future work is the effective knowledge transfer from A1 to A2. A1 plays the role of humans, with full access to all the information, while A2 has only an ambiguous understanding of the surrounding environment from two static video frames after the first phase. In order to describe the video with details that are not included in the initial input, A2 needs to extract useful information from the dialog interactions with A1. Therefore, we will propose a QA-Cooperative network that involves two agents with the ability to process multiple modalities of data. We further propose a cooperative learning method that enables us to jointly train the network with a dynamic dialog history update mechanism. The knowledge gap and transfer process are both experimentally demonstrated.

The novelties of the proposed future work can be summarized as follows: (i) We propose a novel and challenging video description task via two multi-modal dialog agents, whose ultimate goal is for one Agent to describe an unseen video based on the interactive dialog history. This task establishes a more reliable setting by providing implicit information sources to the metamodel. (ii) We propose a QA-Cooperative network and the goal-driven learning method with a dynamic dialog history update mechanism, which helps to effectively transfer knowledge between two agents. (iii) With the proposed network and cooperative learning method, our A2 Agent with limited information can be expected to achieve promising performance, comparable to the strong baseline situation where full ground truth dialog is provided.

\appendix{Background Material}
This paper presents a few simple, yet potentially revolutionary ideas regarding the nature of knowledge. Unfortunately, we find that current habits of cognition can make understanding this paper difficult for some people irrespective of level of formal education, intelligence, and so forth.  This preface is intended to highlight some of the potential blockages to understanding, along with causes, in an attempt to assist readers to potentially benefit from this paper.

We begin by first diving into what may for many appear to be an oxymoronic concept: ``simple yet revolutionary." In general, although this is largely an unstated belief/feeling, it would make much more sense that the term revolutionary would be associated with a certain amount of complexity and/ or dramatic, immediate and obvious impact, e.g., a meteor killing the dinosaurs, Nikola Tesla inventing the induction motor, or Einstein formulating a new theory of space-time. People generally have difficulty gaining an intuitive understanding for small, incremental, simple changes that still have tremendous world changing impact over longer periods of time. 

This particular cognitive blockage, as pointed out by Roberto Unger \citep{unger2015}, is probably at the heart of why society has not yet learned how to continually transform and adapt to changing technological, political, and economic circumstances. Societal changes that are incremental are discounted as insignificant, and those that are dramatic are considered too dangerous, thus locking society into a mode where tectonic political pressures build to monumental levels leading inexorably to world wars and the like. However, for the purpose of understanding this paper, it may help to keep in mind the rice and the chessboard problem \citep{wikipedia2019} which demonstrates how, in an exponential growth situation, a single grain of rice quickly grows to several times the world production of rice \citep{unger2014}.

In order to understand how exponential growth laws apply to systems, such as the AI systems which are the focus of this paper, or societies which we mention only in passing, we remind the reader that technological and knowledge progress at an exponential growth as highlighted in the references (pg. 73 of \citep{korzybski1921}, and \citep{swanson2020}). 

Another major impediment to understanding this paper is what Korzybski termed elementalistic thinking. Elementalistic thinking is the mental habit of attempting to understand systems by focusing primarily on the elements of the systems rather than the interactions between the elements. In complex systems consisting of a large number of elements, understanding individual elements does little to help understand the overall system. For example, in software consisting of millions of lines of code, understanding the individual bits and bytes is of little to no benefit in understanding the nature of the entire software/hardware system.

Korzybski identified many fallacies in our thought processes largely stemming from cultures and educational systems based on modes of thinking originating in Aristotle's time circa 300 BC. While our scientific understanding has progressed by leaps and bounds, we remain largely bound by incorrect legacy beliefs. The metamodel we present in this paper is intended to create intelligent systems that exhibit new levels of autonomous learning and operation. The irony is that to truly understand these concepts, one must be at least partially unhindered by the problems the metamodel solves.

The metamodel is based on the addition of very simple structure to a basic knowledge graph consisting of nodes and links. We add structure to represent distinction relations (i.e. this is not that). These distinction relations we represent as anti-symmetric and form the backbone of our knowledge structure. We then add structure to represent similarity relations. Finally, we formalize Korzybski's structural differential to encode different levels of abstraction \citep{korzybski1949}.

While at the atomic level, these changes are trivial, at the system level, the effects we have seen are revolutionary. The metamodel enables systems to learn and reason without inadvertently mixing levels of abstraction. This actually resolves all of the issues mentioned in this section as ``cognitive blockages."  It should be noted that Korzybski created the field of General Semantics to help humanity evolve to a higher level of cognitive functioning. He spent many years creating a foundation and working with students and educators. He found that younger children were able to fully absorb and adapt to these ideas extremely quickly. Unfortunately for adults, it is a longer process but always successful when there is real interest.

We recommend keeping Baudrillard's idea from Simulacra and Simulation in mind \citep{baudrillard1994}:`` The simulacrum is never that which conceals the truth—it is the truth which conceals that there is none. The simulacrum is true." In the sense that the mental models we humans create define our individual worlds and what is possible within those worlds. When these mental models are incorrect, as they are bound to be at times (e.g. the belief that something cannot be both simple/incremental and revolutionary), it has far reaching implications. At best, our mental models are useful abstractions, e.g. our simplification of the quantum scale world. If we keep this in mind, perhaps we will not attempt to touch the seemingly solid disk mental model we create for a dangerous spinning object like a spinning metal object \citep{korzybski1949}. More importantly, we will continually question whether our views are based on valid abstractions for the current context. As has been pointed out in the machine learning community: ``all models are wrong however some are useful." \citep{box1976}.

\backmatter

\end{document}